\documentclass[runningheads]{llncs}
\usepackage{graphicx}
\usepackage{color}
\usepackage{enumerate}
\usepackage[figuresright]{rotating}
\usepackage{makecell}
\begin{document}
\title{ICDAR 2021 Competition on Scene Video Text Spotting}
%
%
\author{Zhanzhan Cheng\inst{1,2}\thanks{Z. Cheng, J. Lu and B. Zou contributed equally to this competition.} \and
Jing Lu\inst{2}$^\star$ \and
Baorui Zou\inst{3}$^\star$ \and
Shuigeng Zhou\inst{3} \and 
Fei Wu\inst{1} }
\authorrunning{Cheng, Lu and Zou et al.}
%
\institute{Zhejiang University, Hangzhou, China\\
\email{\{11821104,wufei\}@zju.edu.cn} \and
Hikvision Research Institute, Hangzhou, China \\
\email{lujing6@hikvision.com} \and
Fudan University, Shanghai, China\\
\email{\{18210240270,sgzhou\}@fudan.edu.cn}}
\maketitle              

\begin{abstract}
Scene video text spotting (SVTS) is a very important research topic because of many real-life applications. 
However, only a little effort has put to spotting scene video text, in contrast to massive studies of scene text spotting in static images. 
Due to various environmental interferences like motion blur, spotting scene video text becomes very challenging. 
To promote this research area, this competition introduces a new challenge dataset containing 129 video clips from 21 natural scenarios in full annotations. 
The competition containts three tasks, that is, video text detection (Task 1), video text tracking (Task 2) and end-to-end video text spotting (Task3). 
During the competition period (opened on 1st March, 2021 and closed on 11th April, 2021), a total of 24 
teams participated in the three proposed tasks with 
46 
valid submissions, respectively. 
This paper includes dataset descriptions, task definitions, evaluation protocols and results summaries of the ICDAR 2021 on SVTS competition. 
Thanks to the healthy number of teams as well as submissions, we consider that the SVTS competition has been successfully held, drawing much attention from the community and promoting the field research and its development.  
\end{abstract}

\section{Introduction}
Scene video text spotting (SVTS) is a text spotting system for localizing and recognizing text from video flowing, which usually contains multiple modules: video text detection, text tracking and the final recognition. SVTS has become an important research topic due to many real-world applications, including license plate recognition in intelligent transportation system, road sign recognition in advanced driver assistance system or even online handwritten character recognition, to name a few. 
With the rapid development of deep learning techniques, great progress has been made in scene text spotting from static images. However, spotting text from video streams faces more serious challenges than the static OCR tasks in applications. Concretely, SVTS has to cope with various environmental interferences (e.g., camera shaking, motion blur and immediate illumination changing etc.) and meet the real-time response requirement. Therefore, it is necessary to develop efficient and robust SVTS systems for practical applications. 

In recent years, only a little effort has put to spotting scene video text, in contrast to massive studies of text reading in static images. And the studies of SVTS obviously falls behind its increasing applications. This is mainly due to: (1) Though `Text in Videos’ challenges have been recognized since 2013, the dataset is too small (containing only 49 videos from 7 different scenarios), which constrains the research on SVTS. (2) The lack of uniform evaluation metrics and benchmarks, as described in the literature \cite{cheng2019you,cheng2020free}. For example, many methods only evaluate their localization performance on YVT and `Text in Videos' \cite{karatzas2015icdar,karatzas2013icdar}, but few methods pay attention on the end-to-end evaluation. 

Considering the importance of SVTS and the challenges it faces, we propose the ICDAR 2021 competition on SVTS, aiming to draw attention on this problem from the community and promote its research and development. The proposed competition could be of interest to the ICDAR community from two main aspects: 
\begin{itemize}
\item Inherited from LSVTD\cite{cheng2019you,cheng2020free}, the video text dataset is further extended, containing 129 video clips from 21 real-life scenarios. 
Compared to the existing ICDAR video text reading datasets, the extended dataset has some special features and challenges. (1) More accurate annotations compared to existing video text datasets. (2) A general dataset with large range of scenarios, which is collected with different kinds of video cameras: mobile phone cameras in various indoor scenarios (\emph{e.g.}, bookstore and office building) and outdoor street views; HD cameras in traffic and harbor surveillance; and Car-DVR cameras in fast-moving outdoor scenarios (\emph{e.g.}, city road, highway). (3) Some video clips are overwhelming of low-quality images caused by blurring, perspective distortion, rotation, poor illumination or motion inferences (\emph{e.g.} object/camera moving or shaking). To address the potential privacy issue, some sensitive fields (such as person face and vehicle plate license etc.) of the video frames are blurred. The datasets can be an effective complement to the existing ICDAR datasets.
\item Three specific tasks are proposed: video text detection, tracking and the end-to-end recognition. Comprehensive evaluation metrics are used for the three competition tasks, i.e., Recall$_d$, Precision$_d$ and F-score$_d$ \cite{cheng2019you} used for detection, ATA$_t$, MOTA$_t$, MOTP$_t$ 
used for tracking, both sequence-level metrics \cite{cheng2019you} like Recall$_e$, Precision$_e$, F-score$_e$ and the traditional metrics like ATA$_t$, MOTA$_t$, MOTP$_t$ 
used for the end-to-end evaluation. In combination with the extended dataset, it enables wide development, evaluation and enhancement of video text detection, tracking and end-to-end recognition technologies for SVTS. It will help attract wide interests (expected to exceed 50 submits) on SVTS, inspire new insights, ideas and approaches.
\end{itemize}

The competition opened on 1st March, 2021 and closed on 11th April, 2021. 
There are a total of 24 teams participated in the three proposed tasks with {22, 13, 11} valid submissions, respectively. 
This competition report provides the motivation, dataset description, task definition, evaluation metrics, results of submitted methods and their discussion. 
Considering to the large number of teams and submissions, we think that the ICDAR 2021 competition on SVTS is successfully held. 
We hope that the competition draws more attention from the community and further promote the field research and its development.  
\section{Competition Organization}
ICDAR 2021 competition on SVTS is organized by a joint team of Zhejiang University, Hikvision Research Institute and Fudan University. 
The competition make use of Codalab web\footnote[1]{https://competitions.codalab.org/competitions/27667} portal to maintain information of the competition, download links for the datasets, and user interfaces for participants to register and submit their results.  
The schedule of the SVTS competition is as follows:
\begin{itemize}
\item 5 January 2021: Registration is started for competition participants. Training and validation datasets are available for downloads.
\item 1 March 2021: Submissions of all the tasks are open for participants. Test data is released (without ground-truth).
\item 31 March 2021: Registration deadline for competition participants.
\item 11 April 2021: Submissions deadline of all the tasks for participants. 
\end{itemize}
Overall, we received 46 valid submissions from 24 teams from both research communities and industries for the three tasks. 
Note that duplicate submissions are removed.

\section{Dataset}
The dataset has 129 video clips (ranging from several seconds to over 1 minutes long) from 21 real-life scenes. 
It was extended on the basis of LSVTD dataset [2] by addding 15 videos for `harbor surveillance' scenario and 14 videos for `train watch' scenario, for the purpose of addressing video text spotting problem in industrial applications.
\begin{figure}
		\begin{center}
    \includegraphics[width=0.95\linewidth]{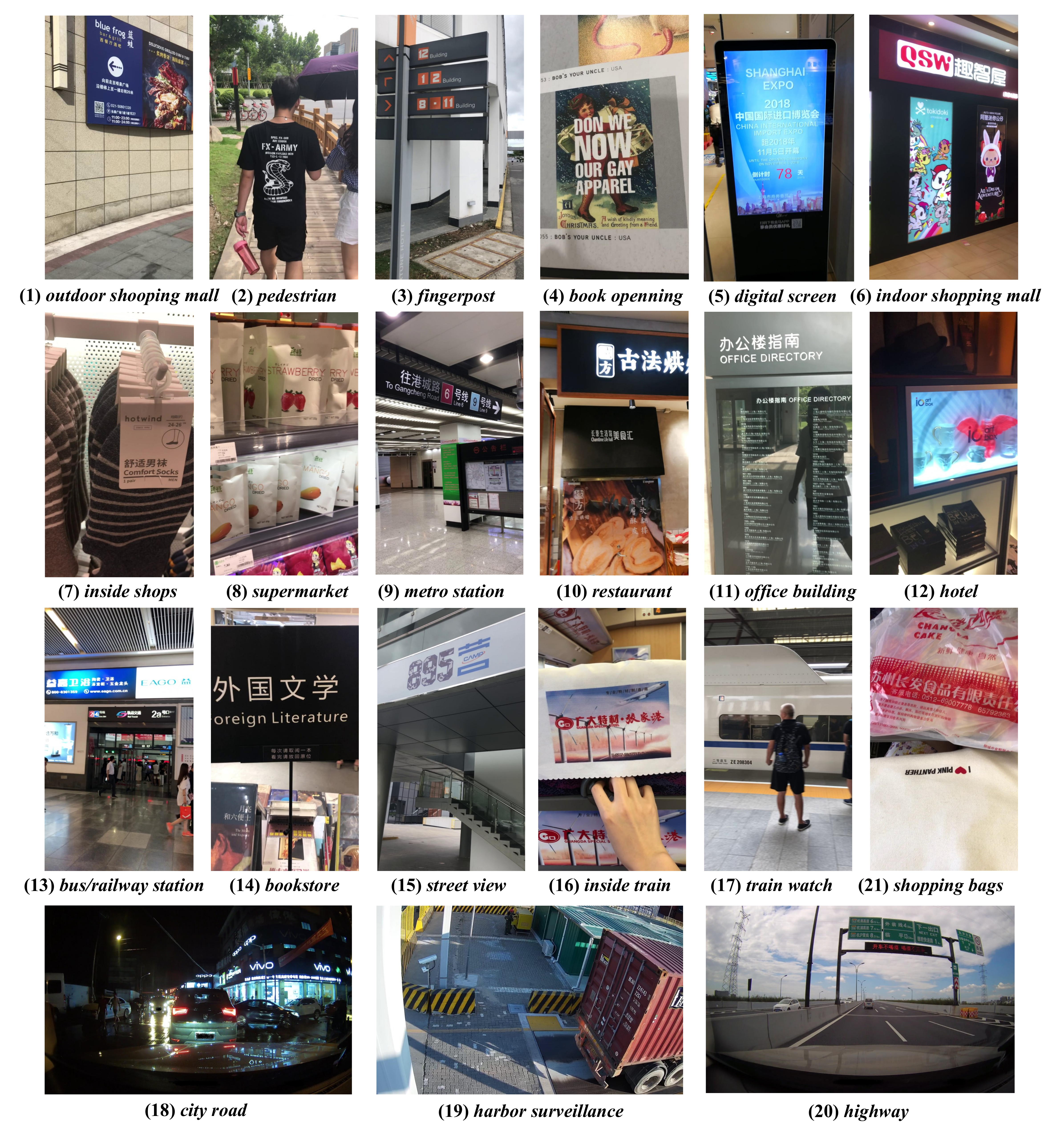}
    \end{center}
    \caption{
    Examples of SVTS for the competition tasks.}
    \label{fig-example}
\end{figure}

\textbf{Characteristics} of the dataset are as follows. 
\begin{itemize}
\item Large scale and diversified scenes. Videos are collected from 21 different scenes, larger than most existing scene video text datasets. 
It contains 13 indoor scenes (reading books, digital screen, indoor shopping mall, inside shops, supermarket, metro station, restaurant, office building, hotel bus/railway station, bookstore, inside train and shopping bags) and 8 outdoor scenes (outdoor shopping mall, pedestrian, fingerposts, street view, train watch, city road, harbor surveillance and highway). 
\item Videos are collected with different kinds of video cameras: mobile phone cameras in various indoor scenarios (e.g. bookstore and office building) and outdoor street views, HD cameras in traffic and harbor surveillance, and Car-DVR cameras in fast-moving outdoor scenarios (e.g. city road, highway). 
\item Different difficulty levels. Hard: videos are overwhelmed by low-quality texts(e.g., blurring, perspective distortion, rotation, poor illumination or even with motion inferences like object/camera moving or shaking). 
Medium: some of the text regions are of low-quality while others are not interfered by artifacts. 
Easy: only a few text regions are polluted in these videos. 
\item Multilingual instances: alphanumeric and non-alphanumeric. 
\end{itemize}

\textbf{Dataset Split}. 
The dataset\footnote[2]{https://competitions.codalab.org/competitions/27667\#learn\_the\_details-datasets}  is divided into training set, validation set and testing set, in which separetely contains 71, 18 and 40 videos. 
The train set contains at least one video from each scenrio.

\textbf{Annotations}. 
The annotation strategy is same to LSVTD \cite{cheng2020free}. For each text region, the annotation items is as follows: 
(1) Polygon coordinate represents text location. 
(2) ID means the unique identification for each text among consecutive frames, i.e., the same text in consecutive frames shares the same ID. 
(3) Language is categorized as Latin and Non-Latin. 
(4) Quality coarsely indicates the quality level of each text region, which can be qualitatively labeled as three quality levels: ‘high’ (recognizable, clear and without interferences), ‘moderate’ (recognizable but polluted with one or several interferences) or ‘low’ (one or more characters are unrecognizable). 
(5) Transcripts mean text string for each text region. 
We parsed videos (ranging from 5 seconds to 1 minute) to frames and then instructed 6 experienced annotation workers to label them, and conducted cross-checking on each text region. 
\section{Tasks}\label{task}
The competition has three tasks: the video text detection, the video text tracking and the end-to-end video text spotting, in which only `alphanumeric text instances' are considered to be evaluated by tools\footnote[3]{https://competitions.codalab.org/competitions/27667\#learn\_the\_details}. 
In future, we intend to release the more challenging multilingual SVTS competition.
\begin{figure}
		\begin{center}
    \includegraphics[width=0.95\linewidth]{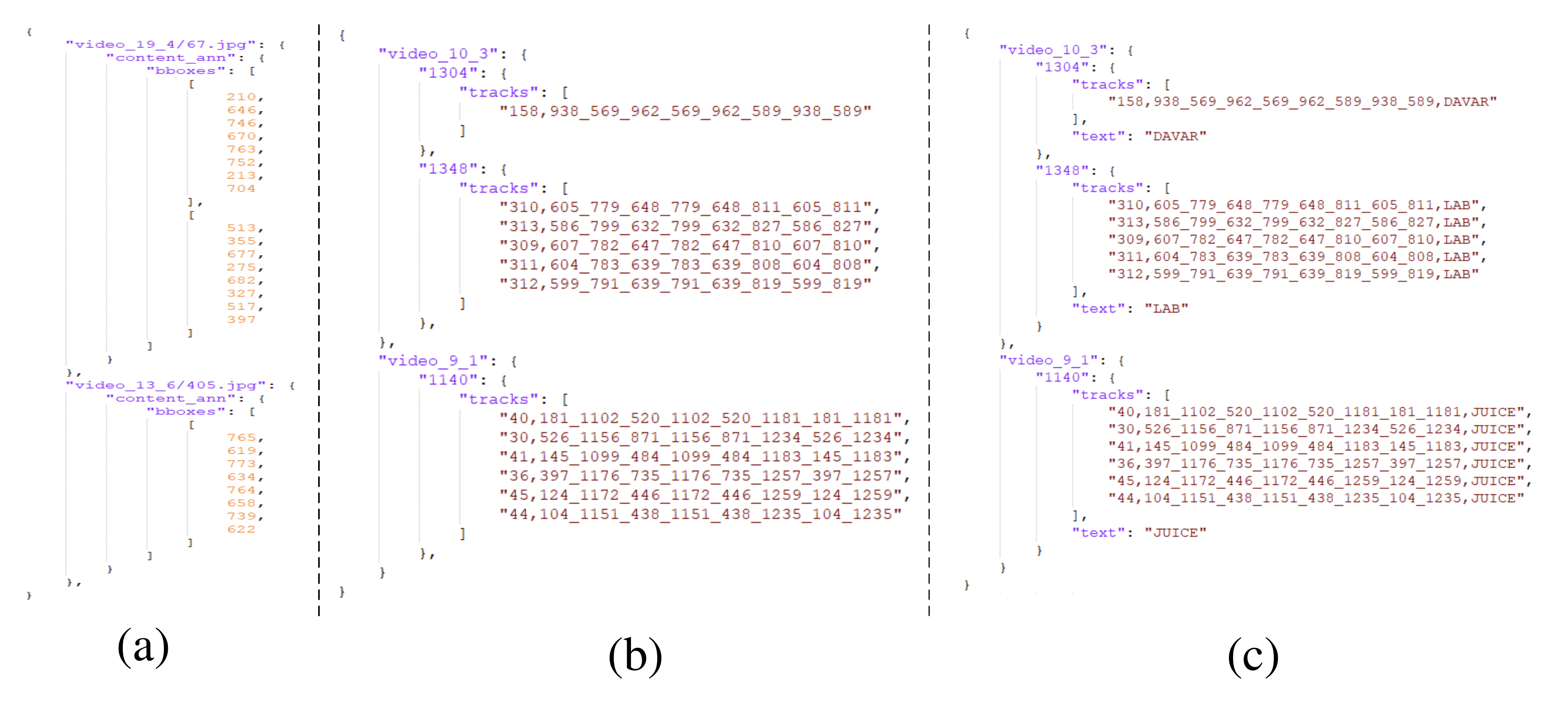}
    \end{center}
    \caption{
    Illustration of the submited JSON files in the three tasks. For each task,  all results are saved in a dict. 
(a) JSON format of video text detection. The field `bboxes' correspond to all the detected bounding boxes in each frame. Each bounding box is represented by 4 points in the clock-wise order.
(b) JSON format of video text tracking, which contains all tracked text sequences in the video. That is, each sequence is represented with a unique ID like `1304' in `Video\_10\_3', and the field `tracks' indicates the tracked bounding boxes in the text sequence. 
(c) JSON format of end-to-end video text spotting, in which a `recognition' word is appended to its bounding box.
}
    \label{fig-json}
\end{figure}

\subsection{Task 1-Video Text Detection}
The task is to obtain the locations of text instances in each frame in terms of their affine bounding boxes. 
Results are evaluated based on the Intersection-over-Union (IoU) with a threshold of 0.5, which is similar to the standard metrics in general object detection like the Pascal VOC challenge. 
Here, Recall$_d$, Precison$_d$ and F-score$_d$ are used as the evaluation metrics. 
The participants are required to prepare a JSON file (named as `detection\_predict.json') containing detection results of all test videos, and then compress and name it as `answer.zip' to upload it. 
The JSON file is illustrated as Figure \ref{fig-json} (a). 

\subsection{Task 2-Video Text Tracking}
This task intends to track all text streams from testing videos. 
Following \cite{karatzas2013icdar}, ATA$_t$, MOTA$_t$ and MOTP$_t$ are used as the evaluation metrics. Note that ATA is selected the main metric because ATA measures the tracking performance over all the text instances.  
Similar to Task 1, the participants are required to prepare a JSON file (named as `track\_predict.json') containing predictions of all test videos, and then compress and name it as `answer.zip' to upload it. 
The JSON file is illustrated as Figure \ref{fig-json} (b). 

\subsection{Task 3-End-to-End Video Text Spotting}
This task aims to evaluate the performance of end-to-end video text spotting. It requires that words should be correctly localized, tracked and recognized simultaneously. 
Concretely, a predicted word is considered as a true positive if and only if its IoU with a ground-truth word is larger than 0.5, and its recognition result is correct at the same time. 

In general, ATA$_e$, MOTA$_e$ and MOTP$_e$ can be used as the tridational evaluation metrics according to the recognition results. 
However, in many real-world applications, the sequence-level spotting results are the most urgently needed for users, while 
it's not what the user cares about for the framewise recognition results.  
Therefore, we propose the sequence-level evaluation protocals to evaluate the end-to-end performance, i.e., Recall$_s$, Precision$_s$, F-score$_s$ as used in \cite{cheng2020free}. 
Here, a predicted text sequence is regarded as a true postive if and only if it satisfies two constraints:
\begin{itemize}
\item The spatial-temporal localisation constraint. The temporal locations of text regions should fall into the interval
between the annotated starting and ending frame. In addition, the given candidate should have a spatial overlap ratio
(over 0.5) with its annotated bounding box. 
\item The recognition constraint: for text sequences satisfying the first constraint, their predicted results should match the corresponding ground truth text transcription. 
\end{itemize}

In order to perform the evaluation, the participants are required to submit a JSON file containing all the predictions for all test videos.
The JSON file should be named as `e2e\_predict.json', and then is submitted by compressing it as `answer.zip'. 
The JSON file is illustrated as Figure \ref{fig-json} (c). 

Some other details that the participants need to pay attention: 
(1) Text areas annotated as `LOW' or `\#\#\#' will not be taken into account for evaluation. 
(2) Words with less than 3 characters are not taken into account for evaluation.
(3) Word recognition evaluation is case-insensitive.  
(4) All the symbols contained in recognition results should be removed before submitted.
(5) The sequence level recognition results should be COMPLETE words.
\section{Submissions}
The submission results are as shown in Table \ref{tab-det}, \ref{tab-track} and \ref{tab-end}. 

\begin{table}
\caption{Video text detection results and rankings after removing the duplicate submissions in task 1. Note that * denotes missing descriptions in affiliations.}
\label{tab-det}
\centering
\scalebox{0.85}{
\begin{tabular}{|c|c|c|c|c|c|}
\hline
User ID &  Rank  & F-score$_d$ & Precision$_d$ & Recall$_d$ & Affiliations\\
\hline
tianqihenhao &  1 & 0.8502 & 0.8561 & 0.8444 & TEG, Tencent\\ \hline
wfeng &  2 & 0.8159 & 0.8787 & 0.7615 & IA, CAS\\ \hline 
\makecell{DXM-DI-AI\\-CV-TEAM} &  3 & 0.7665 & 0.8253 & 0.7155 & DuXiaoman Financial\\ \hline
tangyejun &  4 & 0.7582 & 0.8088 & 0.7136 & *\\ \hline
wangsibo &  5 & 0.7522 & 0.8377 & 0.6825 & *\\ \hline
weijiawu &  6 & 0.7298 & 0.7508 & 0.7098 & Zhejiang University\\ \hline
yeah0110 &  7 & 0.7276 & 0.7314 & 0.7238 & *\\ \hline
BOE\_AIoT\_CTO &  8 & 0.7181 & 0.7133 & 0.7229 & BOE\\ \hline
colorr &  9 & 0.7172 & 0.7101 & 0.7245 & *\\ \hline
qqqyd &  10 & 0.7140 & 0.7045 & 0.7238 & *\\ \hline
yucheng3 &  11 & 0.6749 & 0.8622 & 0.5544 & \makecell{University of Chinese\\Academy of Sciences}\\ \hline
superboy &  12 & 0.6704 & 0.8336 & 0.5607 & *\\ \hline
seunghyun &  13 & 0.6219 & 0.6897 & 0.5663 & NAVER corp\\ \hline
hanquan &  14 & 0.5881 & 0.6252 & 0.5552 & *\\ \hline
sabrina\_lx &  15 & 0.5691 & 0.5463 & 0.5939 & *\\ \hline
gywy &  16 & 0.5640 & 0.5432 & 0.5865 & *\\ \hline
gogogo\_first &  17 & 0.5390 & 0.6521 & 0.4594 & *\\ \hline
SkyRanch &  18 & 0.4766 & 0.4432 & 0.5153 & NUCTech\\ \hline
Steven2045 &  19 & 0.4648 & 0.3614 & 0.6514 & *\\ \hline
enderloong &  20 & 0.3968 & 0.3034 & 0.5732 & *\\ \hline
wuql &  21 & 0.3941 & 0.3039 & 0.5607 & *\\ \hline
AlphaNext &  22 & 0.3373 & 0.2323 & 0.6158 & *\\ 
\hline
\end{tabular}}
\end{table}
\begin{table}
\caption{Video text tracking results and rankings after removing the duplicate submissions in task 2. Note that * denotes missing descriptions in affiliations.}
\label{tab-track}
\centering
\scalebox{0.85}{
\begin{tabular}{|c|c|c|c|c|c|}
\hline
User ID &  Rank & ATA$_t$ & MOTA$_t$ & MOTP$_t$ & Affiliations\\
\hline
tianqihenhao &  1 & 0.5372 & 0.7642 & 0.8286 & TEG, Tencent\\ \hline
\makecell{DXM-DI-AI\\-CV-TEAM} &  2 & 0.4810 & 0.6021 & 0.8017 & DuXiaoman Financial\\ \hline
panda12 &  3 & 0.4636 & 0.7009 & 0.8277 & IA, CAS\\ \hline
lzneu &  4 & 0.3812 & 0.5647 & 0.8198 & *\\ \hline
wangsibo &  5 & 0.3778 & 0.5657 & 0.8200 & *\\ \hline
yucheng3 &  6 & 0.3116 & 0.5605 & 0.8203 &  \makecell{University of Chinese\\Academy of Sciences}\\ \hline
tangyejun & 7 & 0.2998 & 0.5027 & 0.8196 & *\\ \hline
yeah0110 &  8 & 0.2915 & 0.4811 & 0.8218 & *\\ \hline
sabrina\_lx &  9 & 0.2436 & 0.3757 & 0.7667 & *\\ \hline
seunghyun &  10 & 0.1415 & 0.2183 & 0.6949 & NAVER corp\\ \hline
enderloong &  11 & 0.0918 & 0.1820 & 0.7520 & *\\ \hline
tiendv &  12 & 0.0676 & 0.2155 & 0.7439 & \makecell{University of \\Information Technology}\\ \hline
weijiawu &  13 & 0.0186 & 0.1530 & 0.7454 & Zhejiang University\\ 
\hline
\end{tabular}}
\end{table}
\begin{table}
\caption{End-to-end video text spotting results and rankings after removing the duplicate submissions in task 3. Note that * denotes missing descriptions in affiliations.}
\label{tab-end}
\centering
\scalebox{0.85}{
\begin{tabular}{|c|c|c|c|c|c|c|c|c|}
\hline
User ID &  Rank & F-score$_s$ & Precision$_s$ & Recall$_s$ & ATA$_s$ & MOTA$_s$ & MOTP$_s$ & Affiliations\\
\hline
tianqihenhao &  1 & 0.5308 & 0.6655 & 0.4414 & 0.4549 & 0.5913 & 0.8421 &TEG, Tencent\\ \hline
\makecell{DXM-DI-AI\\-CV-TEAM} &  2 & 0.4755 & 0.6435 & 0.3770 & 0.4188 & 0.4960 & 0.8142 &DuXiaoman Financial \\ \hline
panda12 &  3 & 0.4183 & 0.5243 & 0.3479 & 0.3579 & 0.5179 & 0.8427 &IA, CAS\\ \hline
lzneu09 &  4 & 0.3007 &	0.3611 &	0.2576 &	0.2737 &	0.4255&	0.8330 &Northeastern University\\ \hline
yucheng3 &  5 & 0.2964 &	0.3506 &	0.2567 &	0.2711 &	0.4246 &	0.8332 & \makecell{University of Chinese \\Academy of Sciences}\\ \hline
tangyejun &  6 & 0.2284 &	0.2527 &0.2084 &	0.2121 &	0.3676 &	0.8337&* \\ \hline
tiendv &  7 & 0.0813 &	0.1402 &	0.0572 &	0.0802 &	0.0887 &	0.7976 &\makecell{University of\\Information Technology}\\ \hline
enderloong &  8 & 0.0307 &	0.0239 &0.0429 &	0.0357 &	0.0159 &	0.7813 &*\\ \hline
colorr &  9 & 0.0158 &	0.0085 &	0.1225 &	0.0146 &	0.0765 &	0.8498 &*\\ \hline
weijiawu3 &  10 & 0.0077 &	0.0041 &	0.0550 &	0.0088 &-0.1530 &	0.7670 &Zhejiang University\\ \hline
BOE\_AIoT\_CTO &  11 & 0.0000 &	0.0000 &	0.0000 &	0.0000 &	-0.0003 &	0.0000 &BOE\\
\hline
\end{tabular}}
\end{table}
\subsection{Top 3 Submissions in Task 1}
\textbf{Tencent-OCR team} propose the multi-stage text detector by following Cascade Mask R-CNN \cite{2018Cascade} equipped with multiple backbones like HRNet-W48 \cite{2019HRNet}, Res2Net101 \cite{2019Res2Net}, ResNet101 \cite{2016Resnet}, and SENet101 \cite{2018Squeeze}. 
Combined with polygon-NMS \cite{2017Detecting}, the segmentation branch is used to obtain multi-oriented text instances. 
Besides, a CTC \cite{2006Connectionist}-based recognition branch is incorporated into the RCNN stage. 
The whole model is trained in an end-to-end learning pipeline.
To better cope with the challenge of diverse scenes, they employ two approaches in the training phase: (1) Some strong data augmentation strategies are adopted like photometric distortions, random motion blur, random rotation, random crop, and random horizontal flip. (2) Various open-source dataset, e.g., IC13, IC15, IC15 Video, the Latin part of MLT19, COCO-Text, and Synth800k, are involved in the training phase. 
In the inference phase, they make inference by considering to multiple resolutions of 600, 800, 1000, 1333, 1666 and 2000. 
Considering that the image/video quality significantly affects the performance, they design a multi-quality TTA (test time augmentation) approach. With the aid of recognition and tracking results, some detected boxes are removed to achieve higher precision. Finally, four models with distinct backbones are leveraged by ensembling.

\textbf{CASIA\_NLPR\_PAL Team} propose the semantic-aware video text detector, which is an end-to-end trainable video text detector(SAVTD \cite{2021SAVTD}) performing detection and tracking at the same time. In the video text detection task, they adopt Mask R-CNN  \cite{2017Mask} to predict axis-aligned rectangular bounding boxes and the corresponding instance segmentation masks, and then fit a minimum enclosing rotated rectangle to each mask for oriented texts. 

\textbf{DXM-DI-AI-CV-TEAM} use ResNet50 as the backbone. The input scale is 640*640(random crop) and the decoder is Upsample+Conv. Regarding the strategy, the detection results are thoroughly examined by the recognition module in the first place. 
Secondly, a text classifier is trained by utilizing the data from ICDAR2019-LSVT in order to detect and identify the non-text results in the text boxes. The final detection results are then obtained. 

\subsection{Top 3 Submissions in Task 2}
\textbf{Tencent-OCR team} propose the multi-metric text tracking method, which uses 4 different metrics to compare the matching similarity between the current frame detection boxes and the existing text trajectories, i.e., box IoU, text content similarity, box size similarity, and text-geometry neighbor relationship metric. 
The weighted sum of these matching confidence scores are employed as a matching cost between the currently detected box and a tracklet.
Starting from the first frame,they  construct a cost matrix for detected boxes in each frame and existing tracklets. They utilize the Kuhn-Munkres algorithm to obtain matching pairs, and then a grid search is executed to find better parameters. Each box that is not linked with existing tracklets is regarded as a new trajectory. 
They also design a post-processing strategy to reduce ID switches by considering both text regions and recognition results. 
Finally, low-quality trajectories with low text confidence are removed.

\textbf{DXM-DI-AI-CV-TEAM} tackle this task by using a two-fold matching strategy. The ECC \cite{1997Multimodality} algorithm is also utilized to estimate the motion between two video frames and to calculate the estimated bbox for each trajectory.
On the first stage of matching, ResNet34 is employed to extract appearance feature from detection results and estimated bboxes. Next, according to the cosine distance between each feature, a cost matrix is calculated. The matching is then completed in a cascade manner similar to deepsort. On the second stage, the overlap ratio between estimated bbox and  unmatched bouding boxes are used for matching. Strategically, the trajectories with length less than 4 are removed in order to improve the precision on the sequence level.  

\textbf{CASIA\_NLPR\_PAL Team} handles this task via their semantic-aware video text detector. 
Instead of performing text tracking with appearance features extracted from text RoIs directly, they use two fully connected layers to project the roi-features into new ones to get descriptor for each instance, and then matching current frame instance descriptors and previous frame instance descriptors to get current frame tracking identities. In addition to using a new end-to-end trainable method, they also use following strategies to improve the performance. First, in order to train a powerful model, they combine the train set videos and validation set videos for training, in which the base model is pre-training on scene text datasets, such as ArT, MLT. 
Second, since videos in the train set containing different scenes and cover several size ranges, in order to improve the robustness, they also adopt deformable convolution \cite{2017Deformable}. 
In the end, they use ResNet-DCN 50 with multi-scale train/test as the final model. 

\subsection{Top 3 Submissions in Task 3}
\textbf{Tencent-OCR team} develop a method named as Convolutional Transformer for Text Recognition and Correction. 
Two types of networks are leveraged in the recognition stage, including the CTC \cite{2006Connectionist}-based model and the 2D attention sequence-to-sequence model. 
The backbone networks consist of convolutional networks and context extractors. 
They train multiple CNNs including VGGNet, ResNet50, ResNet101, and SEResNeXt50 \cite{2018Squeeze}. 
Then they extract contextual information using BiLSTM, BiGRU, and transformer models. 
For the CTC-based method, they integrate an end-to-end trainable ALBERT \cite{2019ALBERT} as a language model.
The models are pre-trained on 60 million synthetic data samples, and are further fine-tuned on open-source datasets including SVTS, ICDAR-2013, ICDAR-2015, CUTE, IIIT5k, RCTW-2017, LSVT, ReCTS, COCO-Text, RCTW, MLT-2021, and ICPR-2018-MTWI. Data augmentation tricks are also employed, such as Gaussian blur, Gaussian noise, and brightness adjustment. 
In the end-to-end text spotting stage, they predict all detected boxes of a trajectory using different recognition methods. The final text result corresponding to the trajectory is then selected among all recognition results, considering both confidence and character length. Finally, low-quality trajectories whose text result scores are low or whose results contain Chinese characters are removed.

\textbf{DXM-DI-AI-CV-TEAM} achieves the recognition model by using ResNet feature extractor with TPS(Thin Plate Spines) \cite{1989Thin}.  
A relation attention module is employed to capture the dependencies of feature maps and a parallel attention module is used for decoding all characters in parallel. 
The data for training base recognition model contains public datasets such as SynthText, Syn90k, CurvedSynth and SynthText\_Add. To finetune the model, the data with MODERATE and HIGH quality label from the competition datasets are of good use. 
Compared with solutions from other paper to increase the quality scores, this recognition model used the combination of voting and confidence of the results to obtain the final text of the text stream. 

\textbf{CASIA\_NLPR\_PAL Team} handles the  scene text recognition task with sliding convolutional character models. 
For each detected text line, they firstly use a classifier to determine the text direction. Then a sliding-window-based method simultaneously detects and recognizes characters by sliding the text line image with character models, which are learned end-to-end on text line images. The character classifier outputs on the sliding windows are normalized and decoded with CTC-based algorithm. The output classes of the recognizer include all the ASCII and the commonly used Chinese characters. The final adopted model is trained on all competition training and validation data, some publicly released data sets, and a large number of synthetic samples.
\section{Discussion}
In the video text detection task, most participants employ the semantic-based Mask R-CNN framework to capture regular and irregular text instance. 
Tencent-OCR team achieves the best score in F-score$_d$, Recall$_d$ and Precision$_d$ with multiple backbones (e.g., HRNet, Res2Net and SENet, and so on) and model ensembles.
Data augmentation strategies, rich opensource data and multi-scale train/test strategies are useful and important for getting better results.
Besides, many participants employ the end-to-end trainable learning frameworks, which is an obvious research tendency in video text detection.

For video text tracking task, most methods focus on the trajectory estimation by calculating a cost matrix refering to the extracted appearance features. After that, the matching algorithms are employed to generate the final text streams. 
Tencent-OCR team achieves the best score in ATA$_t$, MOTA$_t$ and MOTP$_t$ with multiple metrics (e.g., box IoU, text content similarity, box size similarity and text-geometry neighbor relationship) which are further integrated as the final matching score. 
Besides, some post-processing strategies like removing low-quality text instance are also important for achieve better results. 

For the final text recognition in Task 3, many methods first attempt to train a general model by using various public datasets (e.g., SynthText, Syn90k), and then further finetuned on the released training dataset with MODERATE and HIGH quality. 
Different teams employ different recognition decoders including the CTC-based, 2D attention-based or even the transformer based decoder. 
Tencent-OCR team achieve the 1-st rank in F-score$_e$, Recall$_e$ and Precision$_e$ with various data augmentation, opensource datasets, network backbones and model ensembles.

From the performance of the three tasks, we find that the first two submitted results on detection achieve F-score$_d$ of more than 0.8. 
It indicates that the existing approaches for general text detection are performing well for single video frames. 
While in Task 2 and 3, most submitted methods obtain relatively low scores (less than 0.55) because the video text spotting is very challenging due to various environment interferences.  
Therefore, there are still large space for improvement for the important research topic. 
We also note that, many top ranking methods use ensamble of multiple backbones or metrics to improive performance, and some methods use a wide range of opensource datasets for training a good pre-train model. 
Besides, the end-to-end trainable framework for video text spotting becomes a obvious research trendency. 
Most of submitted methods use different ideas and strategies, and we expect more innovation approaches will be proposed after this competition.  

\section{Conclusion}
This paper summarizes the organization and results of ICDAR 2021 competition on SVTS, detailed on the Codalab website. 
Large scale video text dataset was collected and annotated in full annotations, respectively, containing 21 different scenes.  There have been a number of 24 teams participating in the three tasks and 46 valid submissions in total, which have shown great interest from both research communities and industries. 
Submitted results has shown the abilities of state-of-the-art video text spotting systems. 
On one hand, we intend to keep on maintaining the ICDAR 2021 SVTS competition leaderboard to encourage more participants to submit and improve their results. 
On the other hand, we will extend this competition to multilingual competition for further promoting the research community. 

%
%
%
\bibliographystyle{splncs04}
\bibliography{mybib}
%
\end{document}